\documentclass[letterpaper]{article}
\usepackage[preprint]{twocolpreprint}
\usepackage[hyphens]{url}
\usepackage{graphicx}
\usepackage{natbib}
\usepackage{caption}
\DeclareCaptionStyle{ruled}{labelfont=normalfont,labelsep=colon,strut=off}

\usepackage{booktabs}

\usepackage{amsmath}
\usepackage{amssymb}

\title{Dark Matters: VAE-Aware HDR Encoding for SDR-to-HDR Video Reconstruction}
\author{
    Donghoon Cha\textsuperscript{\rm 1},
    Bomseok Park\textsuperscript{\rm 1},
    Hansl von Kwon\textsuperscript{\rm 1},
    Ku DoHyung\textsuperscript{\rm 1},
    Hanwool Sul\textsuperscript{\rm 2}\corresponding
}
\affiliations{
    \textsuperscript{\rm 1}STUDIO FREEWILLUSION Inc.\\
    \textsuperscript{\rm 2}Seoul National University
}

\begin{document}

\maketitle

\begin{abstract}
        SDR-to-HDR reconstruction is an ill-posed problem in which clipped highlights, crushed shadows, and quantization artifacts must be recovered while preserving the original content. Latent diffusion models provide powerful generative priors for restoring missing detail, but their effectiveness is constrained by how HDR signals pass through the pretrained VAE. We find that common log-like encodings suppress low-luminance chromatic differences before generation, producing a dark-side collapse in the frozen-VAE latent. We introduce \textsc{EquivFlow}, a compact, invertible exposure-equivariant encoding that maps exposure changes to additive shifts while retaining a scale-independent chromatic component, without modifying the VAE. Encoding-level tests across pretrained image and video VAEs show that its exposure structure and deep-dark chroma transport persist across the tested hosts. Integrated into SDR-to-HDR video diffusion, \textsc{EquivFlow} delivers consistent chromatic-fidelity gains across the tested settings while retaining competitive reconstruction fidelity.
\end{abstract}

\section{Introduction}

\begin{figure*}[t]
\centering
\includegraphics[width=503.9881bp]{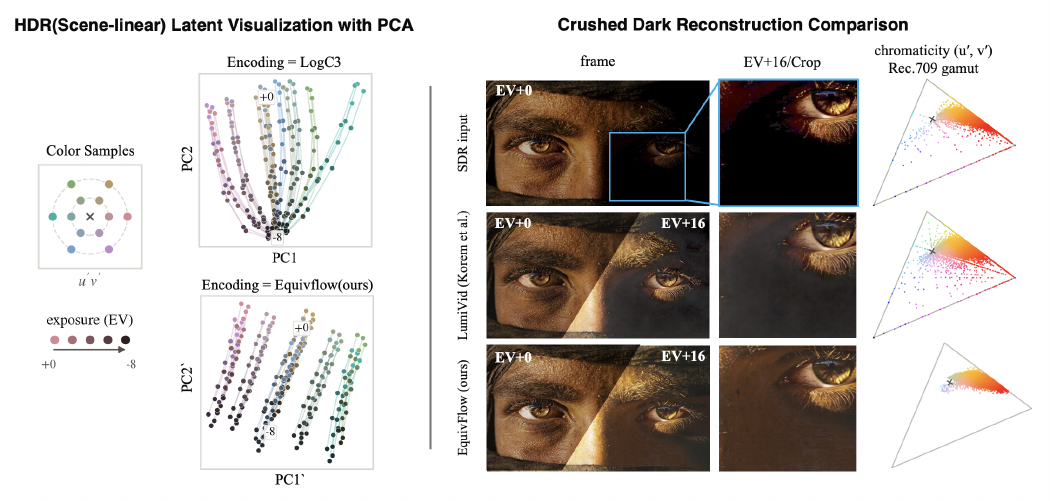}
\caption{Motivation. \textbf{Left:} Frozen-VAE latent PCA for 12 chromaticities over
$0$ to $-8$ EV. LogC3 trajectories collapse to one point at $-8$ EV, while the proposed
exposure-equivariant encoding keeps them parallel and separated. The frozen autoencoder
and color samples are identical, isolating the encoding as the cause. \textbf{Right:} The same
failure in a close-up: the EV$+0$ frame, the crushed region lifted to EV$+16$, and its
CIE 1976 $u'v'$ pixel distribution relative to the Rec.709 gamut.
}
\label{fig:collapse}
\end{figure*}

The growing adoption of high-dynamic-range (HDR) displays and production workflows has
increased the demand for converting existing standard-dynamic-range (SDR) content into
HDR. This demand extends beyond conventional film and video remastering to a range of
applications, including AI-assisted content creation workflows in which image or video
assets are produced in SDR or otherwise limited-dynamic-range form and must be integrated
into HDR production pipelines. SDR-to-HDR reconstruction is therefore becoming
increasingly relevant across these application areas.

Dark-region fidelity is also a practical requirement in HDR mastering. Production quality
control treats crushed blacks as a defect when underexposure removes intended shadow detail,
texture, and depth \citep{netflixCrushedBlacks}. At the signal level, perceptual HDR coding
allocates precision according to luminance-dependent visibility thresholds so that
quantization artifacts remain uniformly small \citep{miller2013pq}. These
practices motivate preserving not only shadow luminance but also low-luminance chromatic
evidence before it enters the pretrained VAE.

Inverse tone mapping is an inherently ill-posed, one-to-many problem. A single SDR
observation may correspond to multiple plausible HDR signals because highlight clipping
removes intensity variation above the SDR range, while shadow crushing, quantization, and
banding suppress subtle luminance and chromatic differences in dark regions. Deterministic
tone expansion alone cannot uniquely recover this missing information. A successful
SDR-to-HDR method must instead infer plausible detail in ambiguous regions while
preserving the spatial and chromatic evidence that remains in the input.

Recent diffusion-based SDR-to-HDR methods address this ambiguity using pretrained
generative models. Before generation, latent implementations encode HDR signals with a
pretrained variational autoencoder (VAE). Recent systems first compress linear HDR with
prescribed log-like transforms---LogC3 in LumiVid and Log-Gamma in DiffHDR---so that it
lies within the pretrained VAE's operating range \citep{lumivid,yu2026diffhdr}. The
representation passed through the VAE can therefore affect which luminance and chromatic
distinctions remain available to the reconstruction model.

\paragraph{Observation 1: latent chromatic structure collapses at low luminance.}
Linear HDR signals are poorly matched to the bounded input range of pretrained VAEs;
Figure~\ref{fig:collapse} qualitatively illustrates how log-like chromatic trajectories
contract in the frozen-VAE latent space at low exposure. Table~\ref{tab:v4-loglike}
quantifies this behavior: LogC3 and Log-Gamma match or exceed \textsc{EquivFlow} in
frozen-VAE round-trip reconstruction, yet retain less linearly recoverable chroma at
\(-8\) EV, with LogC3 unrecoverable there. These frozen-VAE representation and round-trip
measurements diagnose what reaches the generative model; they are not downstream SDR-to-HDR
reconstruction results.

\begin{table}[b]
\centering
\normalsize
\setlength{\tabcolsep}{1.4pt}
\begin{tabular}{@{}lrrrrr@{}}
\toprule
encoding & recon & \(R^2_{u'v'}(0)\) & \(R^2_{u'v'}(-8)\) & surv.\({}_{-8}\) & surv.\({}_{-10}\) \\
\midrule
LogC3 & 35.00 & .481 & -1.133 & -2.354 & -3.410 \\
Log-Gamma & \textbf{36.07} & .738 & .421 & .571 & .391 \\
\textsc{EquivFlow} & 34.38 & \textbf{.801} & \textbf{.883} & \textbf{1.102} & \textbf{1.105} \\
\bottomrule
\end{tabular}
\caption{Frozen-VAE chroma retention for published encodings (no fitted head). \(R^2_{u'v'}\) measures linear chroma recoverability; surv.\({}_{-k}\) is retention after \(k\) stops of darkening (\(1.0\) = unchanged).}
\label{tab:v4-loglike}
\end{table}

\paragraph{Observation 2: exposure changes scale, not chromaticity.}
In linear HDR, multiplicative exposure changes overall scale while leaving chromaticity
unchanged. This suggests an encoding in which exposure becomes translation along a fixed
code-space direction and a transverse chromatic component remains scale-independent. In
our construction, this relation is exact above the luminance toe when both exposure
endpoints remain above it.

These observations motivate \textsc{EquivFlow}, an invertible exposure-equivariant
encoding optimized through the frozen target VAE. It adapts the representation to the
VAE's operating geometry without modifying the pretrained generative backbone.

\paragraph{Our contributions are:}
\emph{(i)} We identify and measure the loss of linearly recoverable chromatic structure in
dark regions after frozen-VAE encoding, and show that the failure is asymmetric across
exposure.
\emph{(ii)} We derive \textsc{EquivFlow}, an invertible, \(13\)-parameter
exposure-equivariant encoding with an analytic inverse, fitted through frozen-VAE
round-trip measurements, and examine its structural behavior across multiple pretrained
video VAEs.
\emph{(iii)} We integrate \textsc{EquivFlow} into SDR-to-HDR video diffusion and evaluate
it against established baselines, obtaining consistent chromatic-fidelity gains across
the tested settings while retaining competitive reconstruction fidelity.

\section{Related Work}

\begin{figure*}[t]
\centering
\includegraphics[width=504.0070bp]{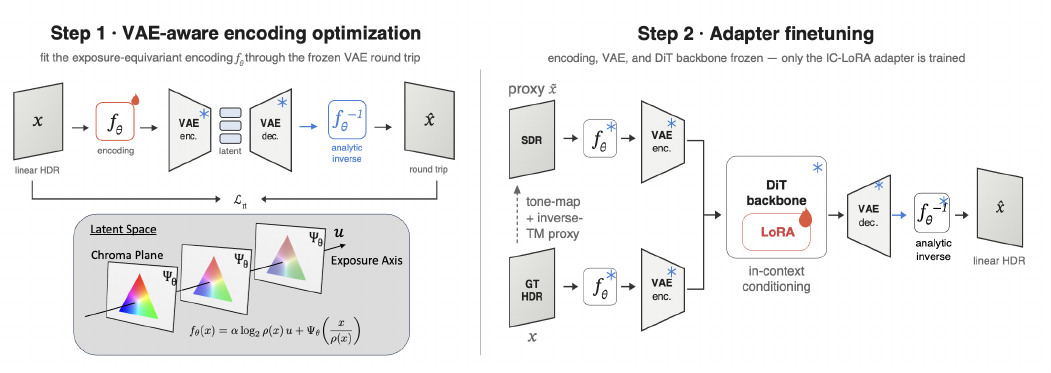}
\caption{Method overview. The display-referred SDR observation is mapped to a
scene-linear proxy by a fixed preprocessing convention, so observation and HDR target
share one scene-linear coordinate system; the invertible exposure-equivariant encoding
\(f_\theta\) then carries that space into the pretrained VAE, and is what the generative
model actually sees. The encoding is fitted first against the frozen VAE, and the adapter
is trained afterwards in the resulting code space. The VAE and the diffusion backbone stay
frozen throughout.}
\label{fig:method-overview}
\end{figure*}

\paragraph{Inverse tone mapping and learning-based SDR-to-HDR.}
Classical iTM expands observed SDR luminance using analytical or spatially adaptive
operators~\citep{banterle2006itmo,rempel2007ldr2hdr}, but cannot model content
missing from the input. Learning-based methods instead predict saturated radiance
with HDRCNN~\citep{eilertsen2017hdrcnn}, regress HDR through multi-scale features
with ExpandNet~\citep{marnerides2018expandnet}, or decompose global mapping, local
enhancement, and highlight generation with HDRTVNet~\citep{chen2021hdrtvnet}.
These supervised feed-forward models motivate generative priors for severely
underdetermined regions.

\paragraph{Diffusion-based SDR-to-HDR reconstruction.}
Diffusion priors offer a way to synthesize plausible content where the SDR input carries no usable evidence. Existing methods incorporate such priors at different stages of reconstruction: Semantic-Aware Diffusion iTM completes saturated regions before merging estimated exposures~\citep{goswami2024saditm}; LEDiff generates and fuses exposure-dependent latents within an LDR latent coordinate system~\citep{wang2024lediff}; and gain-map diffusion predicts an SDR base and a gain map~\citep{guan2025gainmap}.

Many of these diffusion-based approaches build on latent diffusion models, which perform denoising in the latent space of a pretrained autoencoder rather than directly in pixel space~\citep{rombach2022ldm}. This formulation is particularly attractive for SDR-to-HDR reconstruction, as it enables the reuse of strong pretrained image or video priors while operating on compact perceptual representations. Adapting an SDR-pretrained latent diffusion model to HDR, however, also requires mapping HDR signals into a form compatible with its pretrained autoencoder. LumiVid addresses distribution alignment by applying a fixed LogC3~\citep{arri2023logc3} transform before a frozen video VAE, thereby mapping scene-linear HDR signals closer to the pretrained latent distribution~\citep{lumivid}. DiffHDR follows a related design using a fixed Log-Gamma transform with a frozen video VAE~\citep{yu2026diffhdr}. These designs highlight HDR encoding as an important component of latent diffusion pipelines for SDR-to-HDR reconstruction.

\paragraph{Luminance and chrominance.}
Low-light enhancement methods have also emphasized how luminance interacts with color.
HVI introduces a trainable color space that separates intensity from chromatic components
and compresses low-intensity coordinates to suppress noise~\citep{yan2025hvi}.
ICLR further improves luminance--chrominance interaction through dual-stream feature
exchange and a covariance correction loss~\citep{xu2026iclr}. These methods show that
restoration quality depends not only on network capacity, but also on how luminance and
chromatic information are organized before feature extraction. This consideration becomes
particularly important when signals undergo strong dynamic-range compression, since subtle
chromatic distinctions in dark regions may be attenuated before later restoration stages
can exploit them.

\section{Method}

\subsection{Overview}

Our method reconstructs scene-linear HDR video from a display-referred SDR observation
through a frozen latent video diffusion model. It is defined by three choices. The SDR
observation is first mapped to a scene-linear proxy by a fixed preprocessing convention,
so that the observation and HDR target can be expressed in one scene-linear coordinate
system. That space is then carried into the pretrained VAE by an invertible
exposure-equivariant color encoding $f_\theta$, which is what the generative model
actually sees. Here, ``encoding'' denotes an invertible, pointwise reparameterization
of scene-linear color rather than a transformation of scene content. The encoding is
learned first, against the frozen VAE, so that subsequent
adapter training operates in a code space already adapted to the bottleneck it must pass
through.

\subsection{Exposure Equivariance}

The constraint is motivated by what the frozen VAE receives. Log-like encodings can
shrink dark chromatic differences before the VAE sees them
(Figure~\ref{fig:collapse}), leaving the generative prior conditioned on an already
attenuated signal. We therefore seek an encoding in which exposure changes move codes
predictably without reducing chromatic separations.

Exposure acts on scene-linear color by a positive scalar, $x\mapsto sx$. We require the
encoding to turn this multiplicative action into a translation along a fixed unit
direction $u$ of code space:
\begin{equation}
f_\theta(sx)
=
f_\theta(x)
+
\alpha\log_2(s)\,u,
\qquad
\lVert u\rVert_2=1,
\label{eq:equiv}
\end{equation}
with $\alpha$ the code displacement per exposure stop.

Equation~\eqref{eq:equiv} gives exposure and chromaticity separate roles in the code.
Along $u$ the code travels with exposure; transverse to $u$ it is untouched by any global
scaling and therefore carries chromatic information. In particular, code differences are
exposure-invariant, $f_\theta(sx)-f_\theta(sy)=f_\theta(x)-f_\theta(y)$, so chromatic
amplitude does not contract merely because luminance falls.

The requirement is restrictive enough to be characterized. A map satisfies
Eq.~\eqref{eq:equiv} for all $s>0$ if and only if it decomposes as
\begin{equation}
f_\theta(x)
=
\alpha\log_2\rho(x)\,u
+
\Psi_\theta\!\left(\frac{x}{\rho(x)}\right),
\label{eq:encoding_decomposition}
\end{equation}
where $\rho$ is positive and $1$-homogeneous ($\rho(sx)=s\rho(x)$) and $\Psi_\theta$ is
defined on the chromaticity slice $\{x:\rho(x)=1\}$, with its image taken transverse to
the exposure direction $u$. Taking $\Psi_\theta$ to be the restriction of $f_\theta$ to
that slice and applying Eq.~\eqref{eq:equiv} with
$s=\rho(x)$ gives one direction; $1$-homogeneity of $\rho$ gives the other. The
supplement states and proves this formally.

Equation~\eqref{eq:encoding_decomposition} gives a practical template for the encoding.
The function $\rho$ chooses the brightness coordinate, $\Psi_\theta$ encodes the color
remaining after brightness normalization, and $\alpha$ sets how far the code moves per
exposure stop. Since the VAE accepts only a bounded input range, these choices are fixed
as a measured operating point for the target VAE rather than set from color-science
principles alone. The construction gives an invertible representation in which exposure
and color have explicit roles before the representation is fitted to the target VAE.

\subsection{Encoding Parameterization}

Our encoding, \textsc{EquivFlow}, realizes Eq.~\eqref{eq:encoding_decomposition} in closed
form with $13$ fitted parameters. We take $\rho=Y$, the Rec.709 luminance~\citep{itu2015bt709}, which is positive and
$1$-homogeneous, and coordinatize the chromaticity slice by the CIE 1976 coordinates
$(u',v')$~\citep{cie2024uv}, which are invariant under $x\mapsto sx$ by construction. Both roles of
Eq.~\eqref{eq:encoding_decomposition} are then linear, and the encoding is a single
affine map of the feature triple $\phi(x)=\big(\ell_\tau(Y),\,u',\,v'\big)$:
\begin{equation}
f_\theta(x)=B\,\phi(x)+b,
\label{eq:lite}
\end{equation}
whose intensity coordinate is a toe-limited logarithm,
\begin{equation}
\ell_\tau(Y)=
\begin{cases}
\log_2 Y, & Y\geq\tau,\\[2pt]
\log_2\tau+\dfrac{Y/\tau-1}{\ln 2}, & Y<\tau .
\end{cases}
\label{eq:toe}
\end{equation}
The fitted toe $\tau$ is required because $\log_2$ diverges as $Y\to0$ while the code
range is bounded; below it the intensity coordinate continues linearly with matched value
and slope. This keeps the luminance coordinate finite near black without introducing a
discontinuity at the toe.

Above the toe the encoding satisfies Eq.~\eqref{eq:equiv} exactly rather than
approximately: $u'$ and $v'$ are unchanged by scaling and
$\ell_\tau(sY)=\ell_\tau(Y)+\log_2 s$, so $f_\theta(sx)=f_\theta(x)+\log_2(s)\,Be_1$. The
flow direction and the per-stop budget are therefore read off the first column of $B$,
\begin{equation}
u=\frac{Be_1}{\lVert Be_1\rVert_2},
\qquad
\alpha=\lVert Be_1\rVert_2 ,
\label{eq:flow_from_B}
\end{equation}
so the exposure direction and per-stop displacement are determined by the first column of
$B$. Inversion needs no learned decoder: solving the affine system gives
$\phi=B^{-1}(z-b)$, from which $Y=\ell_\tau^{-1}(\phi_1)$ and
$(Y,u',v')$ maps back to linear Rec.709 in closed form, followed by the nonnegativity
projection. The $13$ parameters are the $9$ entries of $B$, the $3$ offsets $b$, and
the toe $\tau$.

\subsection{VAE-Aware Optimization}

\begin{table*}[t]
\centering
\small
\setlength{\tabcolsep}{4.0pt}
\begin{tabular*}{\textwidth}{@{\extracolsep{\fill}}l||cc|cc||cc|cc||cc|cc@{}}
\toprule
method & \multicolumn{4}{c}{ARRI (40)} & \multicolumn{4}{c}{HDRTV4K (40)} & \multicolumn{4}{c}{UPIQ (22)} \\
\cmidrule(lr){2-5}\cmidrule(lr){6-9}\cmidrule(lr){10-13}
& \multicolumn{2}{c}{PU21-PSNR $\uparrow$} & \multicolumn{2}{c}{$\Delta C_TC_P$ $\downarrow$} & \multicolumn{2}{c}{PU21-PSNR $\uparrow$} & \multicolumn{2}{c}{$\Delta C_TC_P$ $\downarrow$} & \multicolumn{2}{c}{PU21-PSNR $\uparrow$} & \multicolumn{2}{c}{$\Delta C_TC_P$ $\downarrow$} \\
\cmidrule(lr){2-3}\cmidrule(lr){4-5}\cmidrule(lr){6-7}\cmidrule(lr){8-9}\cmidrule(lr){10-11}\cmidrule(lr){12-13}
\textit{TM method} & rein. & aces & rein. & aces & rein. & aces & rein. & aces & rein. & aces & rein. & aces \\
\midrule
\textsc{EquivFlow} & \textbf{39.99} & 26.11 & \textbf{5.51} & \textbf{5.97} & \textbf{33.12} & 26.16 & \textbf{8.67} & \textbf{10.23} & \textbf{41.77} & 27.61 & \textbf{3.02} & \textbf{4.27} \\
LogC3 & 39.70 & \textbf{26.33} & 5.52 & 6.23 & 31.43 & \textbf{27.44} & 9.73 & 11.95 & 39.28 & \textbf{28.63} & 3.93 & 5.59 \\
\midrule
LumiVid & 36.26 & 25.98 & 6.87 & 8.18 & 32.36 & 27.25 & 10.86 & 14.54 & 36.01 & 26.94 & 6.18 & 10.30 \\
HDRTVNet++ & 26.32 & 20.61 & 6.05 & 6.23 & \textemdash & \textemdash & \textemdash & \textemdash & 28.07 & 22.48 & 4.96 & 6.19 \\
LEDiff & 18.37 & 18.74 & 12.25 & 12.29 & 18.88 & 19.12 & 18.87 & 19.41 & 20.10 & 20.64 & 10.29 & 10.74 \\
\bottomrule
\end{tabular*}
\caption{PU21-PSNR and the chromatic $\Delta E_{\mathrm{ITP}}$ term $\Delta C_TC_P$ on a
frozen 102-clip test set --- the first 40 ARRI crops (10 parents), the first 40 HDRTV4K
clips (19 parents), and all 22 UPIQ parents --- frozen before any aggregation was
computed, under the training curve (\emph{rein.}) and one unseen curve (\emph{aces});
the HDR ground truth, clip list, per-clip normalization and crop are held fixed while
the SDR tone curve moves. \textsc{EquivFlow} holds the lowest $\Delta C_TC_P$ in all
six chromatic columns and the best PU21-PSNR under the training curve on all three
sets; under the unseen curve the aggregate lead passes to LogC3. $\Delta C_TC_P$ is
defined on PQ-encoded ICtCp and is not exposure-invariant; the exposure-invariant
$\Delta u'v'$ reading is Table~\ref{tab:v5-chroma}. Parent-weighted means; clip counts
in parentheses; best per column in bold. Upper block: the matched internal pair; lower
block: external systems. \textemdash: withheld, same-domain with HDRTVNet++'s training
corpus and never allowed to win a column.}
\label{tab:v5-main}
\end{table*}

Analytic invertibility does not imply survival through the pretrained bottleneck. The
encoder is frozen and has its own operating geometry, so theclosed-form parameters of
Eq.~\eqref{eq:lite} are fitted against that geometry rather than from color-science
principles alone. For the encoding $f_\theta$, the relevant measurement is the frozen VAE
round trip
\begin{equation}
x
\rightarrow
f_\theta(x)
\rightarrow
D\left(E(f_\theta(x))\right)
\rightarrow
f_\theta^{-1}\left(D(E(f_\theta(x)))\right),
\label{eq:vae_roundtrip}
\end{equation}
where $E$ and $D$ are the frozen encoder and decoder and the final inverse is followed by
the nonnegativity projection.

Let $\hat{x}_\theta$ denote the output of this round trip. For \textsc{EquivFlow}, we
score VAE compatibility by the round-trip reconstruction loss
\begin{align}
\mathcal{L}_{\mathrm{rt}}
=&
\left\lVert \log(1+\hat{x}_\theta)-\log(1+x)\right\rVert_2^2
\nonumber\\
&+
\lambda_{uv}\left\lVert m(x)\big(q(\hat{x}_\theta)-q(x)\big)\right\rVert_2^2 .
\label{eq:roundtrip_loss}
\end{align}
Here $q(x)=(u'(x),v'(x))$ and $m(x)$ masks pixels above the dark-noise floor. The
encoding is also required to keep codes in the VAE input range before hard clipping.

The \textsc{EquivFlow} encoding is an exactly exposure-equivariant closed form with an
analytic inverse, fitted by frozen-VAE round-trip measurements before any task-level
diffusion evaluation.

The fitted parameters belong to the VAE they were fitted against; how far one parameter
set transfers to other frozen bottlenecks is measured in the experiments rather than
assumed.

\subsection{SDR-to-HDR Reconstruction}

After fitting, the encoding is fixed. For supervised training and evaluation, each
scene-linear HDR target $x$ is rendered to a display-referred SDR observation by a fixed
recipe: normalization, extended Reinhard tone mapping, per-channel clipping, the sRGB OETF~\citep{iec1999srgb}, and $8$-bit
quantization. The SDR observation is then mapped to a scene-linear
proxy $\tilde{x}$ using a fixed Reinhard inverse-tone-mapping preprocessing convention.
This proxy defines the conditioning coordinate used by the model, while clipped,
crushed, and quantized content remains underdetermined. Because the preprocessing
contract may affect performance, the experiments additionally evaluate tone-mapper
variants rendered with Reinhard~\citep{reinhard2002photographic} and ACES-like~\citep{narkowicz2016aces} curves.

Both $x$ and $\tilde{x}$ are mapped through the same encoding $f_\theta$ and encoded
online by the frozen VAE. The target latent serves as the diffusion target, while the
proxy latent is supplied as in-context conditioning.

At inference, the conditioned diffusion model predicts an HDR encoding latent. The frozen
VAE decoder maps this latent to encoding space, after which the analytic inverse of
$f_\theta$ maps the result to linear HDR:
\begin{equation}
\hat{x}
=
f_\theta^{-1}\left(D(\hat{h})\right).
\label{eq:inference}
\end{equation}
\paragraph{Diffusion backbone and IC-LoRA adaptation.}
The reconstruction model uses an in-context LoRA conditioning scheme
\citep{huang2024iclora}. In our implementation, the SDR condition latent is formed from
the scene-linear proxy after the same color encoding $f_\theta$, rather than from raw
SDR. It is prepended as clean reference tokens to the noised HDR target-token sequence;
reference tokens receive zero timestep and are excluded from the denoising loss, which is
evaluated only on the target portion. Only the low-rank adapter~\citep{hu2022lora} is optimized; the VAE,
diffusion backbone, and text encoder remain frozen.

\begin{figure}[t]
\centering
\includegraphics[width=214.6468bp]{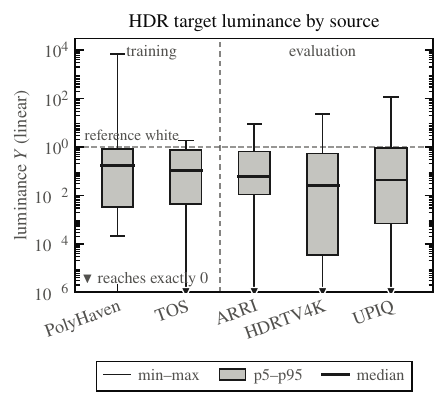}
\caption{HDR target luminance by source (log axis, pooled over sampled pixels): line and
caps span minimum to maximum, the bar the 5th--95th percentile, the dot the median, and the dashed line marks
reference white. Target, not the SDR the model receives.}
\label{fig:source_range}
\end{figure}

\section{Experiments}

\begin{figure*}[t]
\centering
\includegraphics[width=504.0487bp]{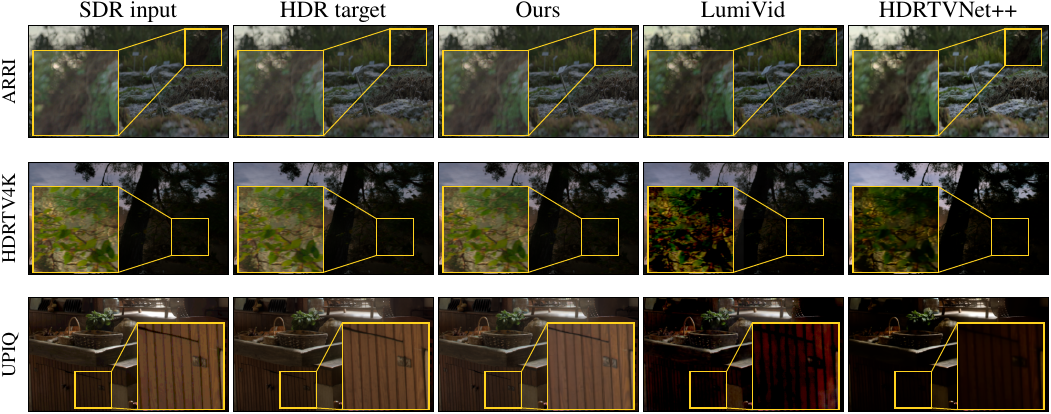}
\caption{Qualitative comparison at frame 24.  The inset covers the darkest chroma-carrying ground-truth region;
each row carries a display gain applied identically across columns.}
\label{fig:v5-qualitative}
\end{figure*}

\subsection{Datasets and Setup}

\textbf{Training corpus.}
All internal arms are trained on one corpus of 5{,}780 crops at
\(960\times544\) and 49 frames, drawn from 1{,}015 parent sequences: 215
windows from 94 \emph{Tears of Steel}~\citep{blender2012tearsofsteel} shots (2{,}580 crops) and 800 clips
rendered from PolyHaven~\citep{polyhaven} HDRI panoramas (3{,}200 crops). Splits are
parent-disjoint, shot-level for \emph{Tears of Steel} and asset-level for
PolyHaven.

These are fixed derivatives of the cited public sources, not a newly introduced dataset or benchmark.

\textbf{Training inputs.}
Training SDR inputs use the fixed Reinhard-based rendering protocol described in the
Method section, followed by the same camera-style degradations for every internal arm:
contrast scaling, band-selective blur, 8-bit quantization, and codec compression.
Evaluation additionally includes ACES-like renderings.

\textbf{Evaluation sets.}
We evaluate on ARRI's official ALEXA Mini ProRes 4444 XQ sample reel
M001R00H~\citep{arriSampleFootage}, HDRTV4K~\citep{guo2023hdrtv4k}, and
UPIQ~\citep{mikhailiuk2021upiq}, decoding every prediction to common linear
Rec.\ 709~\citep{itu2015bt709} before scoring. ARRI contains 176 crops from 44
cinema-camera parents and is the brightest set. HDRTV4K contains 78 clips
from 36 parents of mixed video and static content, including material from
the source used by HDRTVNet++. UPIQ contains one 49-frame static clip for each
of 22 calibrated image parents, so temporal metrics are undefined there.
The main table reports a 102-clip test subset frozen before any aggregation
was computed: the first 40 ARRI crops (10 parents), the first 40 HDRTV4K
clips (19 parents), and all 22 UPIQ parents. The chromatic-decomposition
panel is computed on the full HDRTV4K--UPIQ pool.

\textbf{Baselines.}
The internal rows share the training split above, the frozen LTX-2.3 22B
backbone and VAE~\citep{hacohen2024ltxvideo,lightricks2026ltx23}, IC-LoRA adapter
\citep{huang2024iclora}, inference settings, and scorer. We report the protocol
from the Method section: canonical Reinhard inputs and one unseen tone mapper,
ACES-like. LogC3 is the matched internal baseline: it is
retrained on the same corpus and degradation protocol with the same backbone,
conditioning method, and scorer. The two internal arms differ in the HDR
representation and in its paired reference contract: \textsc{EquivFlow}
conditions on the inverse-tone-mapped scene-linear proxy of the Method section,
which is part of the contribution, while LogC3 is trained under raw-SDR
conditioning, the official contract of LumiVid.

LumiVid is reported separately because it is the closest architectural
predecessor: a pretrained video diffusion model adapted for SDR-to-HDR through
a fixed LogC3 interface. HDRTVNet++~\citep{chen2023hdrtvnetpp} and
LEDiff~\citep{wang2024lediff} are external system baselines that differ in
architecture, training data, and budget. LEDiff uses its official preprocessing
and one method-level radiometric gain fitted on a frozen 24-clip training
subset, without access to test images or per-clip ground truth.

\subsection{Main Results}

\textbf{Quantitative results.}
Table~\ref{tab:v5-main} evaluates the frozen test subset under the training
Reinhard curve and an unseen ACES-like curve, with the target and evaluated
clips held fixed. PU21-PSNR~\citep{mantiuk2021pu21} measures aggregate HDR
reconstruction, while \(\Delta C_{T}C_{P}\), the chromatic component of
\(\Delta E_{\mathrm{ITP}}\)~\citep{itu2019bt2124}, measures perceptual chromatic
error at absolute luminance. Under the training curve, \textsc{EquivFlow} leads
the matched LogC3 control on both quantities across all three sets. Under the
unseen curve, it retains the chromatic lead throughout, while LogC3 takes the
aggregate PU21-PSNR lead. The consistent result is therefore improved chromatic
fidelity across tone curves; aggregate reconstruction depends more strongly on
the SDR rendering. Parent-level tests support the matched-pair chromatic difference
on UPIQ under the training curve and on HDRTV4K and UPIQ under the unseen curve;
the smaller ARRI subset does not separate under the training curve. Remaining
preregistered metrics and full statistical details are reported in the supplement.

\textbf{Baseline comparison.}
LumiVid is the closest architectural predecessor and therefore the most direct
external comparison. \textsc{EquivFlow} has lower chromatic error in every
column and higher aggregate fidelity in all but the HDRTV4K ACES-like case.
It also leads LEDiff and HDRTVNet++ on both reported quantities wherever a
comparison is available; HDRTVNet++ is withheld on HDRTV4K because that set is
same-domain. These remain system-level comparisons because architecture,
training data, radiometric convention, and budget differ.

\begin{table}[t]
\centering
\small
\setlength{\tabcolsep}{3.2pt}
\begin{tabular}{@{}lcc cc@{}}
\toprule
& \multicolumn{2}{c}{reinhard} & \multicolumn{2}{c}{aces-like} \\
\cmidrule(lr){2-3}\cmidrule(lr){4-5}
metric \(\downarrow\) & Ours & LogC3 & Ours & LogC3 \\
\midrule
chroma L1 & \textbf{.0431} & .0539 & \textbf{.0525} & .0656 \\
\(\Delta u'v'\) & \textbf{.0140} & .0173 & \textbf{.0170} & .0247 \\
\(\Delta C_{T}C_{P}\) & \textbf{6.37} & 7.37 & \textbf{7.98} & 9.47 \\
\bottomrule
\end{tabular}
\caption{Chromatic error of the internal pair on the pooled HDRTV4K--UPIQ panel (58 parents), training curve and one unseen curve. Lower is better; parent-weighted; best per pair in bold. All six paired differences favor Ours and survive Holm correction (largest corrected \(p=4.8\times10^{-6}\)). ARRI appears in Table~\ref{tab:v5-main} only; nested \(\Delta u'v'\) variants and per-metric intervals: supplement.}
\label{tab:v5-chroma}
\end{table}

\textbf{Chromatic decomposition.}
To test whether the main-table chromatic trend depends on one metric,
Table~\ref{tab:v5-chroma} evaluates the matched pair using raw chroma amplitude,
exposure-invariant CIE 1976 chromaticity, and PQ-encoded
ICtCp~\citep{itu2018bt2100}. Every metric favors \textsc{EquivFlow} under both
tone curves, with a larger margin under the unseen curve. This consistency
supports the chromatic-fidelity reading across the tested metrics and curves.

\textbf{Qualitative results.}
Figure~\ref{fig:v5-qualitative} shows the highest-margin
\textsc{EquivFlow}--LumiVid example from each evaluation set, so the rows
show selected successes rather than typical cases. In the magnified
dark regions, \textsc{EquivFlow} remains closer to the target's color and local
structure, while the comparison outputs show stronger saturation, hue, or
shadow-level deviations. These examples complement, but do not replace, the
dataset-level quantitative comparisons above.

\subsection{Representation Analysis}

Table~\ref{tab:v4-loglike} tests representation behavior before diffusion by
passing each published encoding through the same frozen VAE without a fitted
head. The log-like encodings reconstruct well but lose linearly recoverable
chroma as exposure falls, whereas \textsc{EquivFlow} preserves it; reconstruction
quality is therefore not a proxy for dark-exposure chroma retention.
This divergence is consistent with exposure--chroma coupling and matches the
downstream widening of the chromatic margin under the unseen tone curve
(Tables~\ref{tab:v5-main} and~\ref{tab:v5-chroma}). We next examine the
supporting design choices.

\textbf{Cross-VAE transfer test.}
\textsc{EquivFlow}'s exposure structure is tested on frozen LTX, Stable
Diffusion~2.1~\citep{rombach2022ldm,sd21modelcard}, Wan~2.1 T2V-1.3B~\citep{teamwan2025},
and Wan~2.2 TI2V-5B~\citep{wan2025wan22}
VAEs; only LTX was used during fitting. On the three foreign hosts, the
exposure-linked latent displacement remains nearly
linear and chromaticity readout stays above chance at deep exposure. All three
foreign hosts pass the frozen roundtrip non-collapse rule in that regime.

Portability here concerns the exposure structure, not the latent coordinates
themselves: an EV-0-frozen readout degrades along each host's exposure
trajectory, and absolute round-trip quality varies substantially across hosts.

\begin{table}[t]
\centering
\normalsize
\setlength{\tabcolsep}{2.1pt}
\begin{tabular}{@{}lcccc@{}}
\toprule
VAE & \(R^2_{\beta}\uparrow\) & grid \(0\!\rightarrow\!-8\uparrow\)
 & \(\Delta u'v'_{-8}\downarrow\) & /LTX \\
\midrule
LTX       & 0.9990 & \(0.979\!\rightarrow\!0.979\)      & 0.01536 & 1.000 \\
SD~2.1    & 0.9927 & \(0.667\!\rightarrow\!0.812\)      & 0.00756 & 0.492 \\
Wan~2.1   & 0.9977 & \(0.958\!\rightarrow\!0.917\)      & 0.00824 & 0.536 \\
Wan~2.2   & 0.9976 & \(0.979\!\rightarrow\!0.854\)      & 0.00793 & 0.516 \\
\bottomrule
\end{tabular}
\caption{Training-free portability of \textsc{EquivFlow}, measured on this draft's carrier. Grid accuracy uses a host- and exposure-specific readout; \(\Delta u'v'\) is frozen-VAE round-trip error and /LTX is that error relative to the LTX host.}
\label{tab:v5-crossvae}
\end{table}

\section{Conclusion}

We introduced \textsc{EquivFlow}, a VAE-aware HDR encoding motivated by the loss of
recoverable chromatic structure under darkening in common log-like representations.
Its compact, invertible, pointwise construction maps exposure changes to code-space
translations while preserving a scale-independent chromatic component, without modifying
the pretrained VAE or diffusion backbone.
Representation analyses across pretrained image and video VAEs confirm the persistence of
this structure, and integration into SDR-to-HDR video diffusion yields consistent
chromatic-fidelity gains across the tested datasets and tone curves while retaining
competitive reconstruction fidelity.
Together, these results establish VAE-aware encoding as an effective interface for adapting
pretrained latent generative models to HDR reconstruction. Future work will examine this
design across broader VAE families and reconstruction settings.

\bibliography{refs}

\clearpage
\appendix
\setcounter{equation}{0}\setcounter{table}{0}\setcounter{figure}{0}
\begin{center}{\Large\bf Supplementary Material}\end{center}
\vspace{0.5em}

\section{Exposure-Equivariant Encodings}

This section states the general characterization used in the main paper and
then verifies the properties of the realized encoding. Let
$f:\mathbb{R}_{>0}^{3}\rightarrow\mathbb{R}^{d}$ be a code map, let
$u\in\mathbb{R}^{d}$ be a unit vector, and let $\alpha>0$. We say that $f$ is
exposure-equivariant along $u$ if, for every color $x\in\mathbb{R}_{>0}^{3}$
and every exposure scale $s>0$,
\begin{equation}
f(sx)=f(x)+\alpha\log_2(s)u.
\label{eq:supp-equiv}
\end{equation}

\paragraph{Proposition 1 (characterization).}
Let $\rho:\mathbb{R}_{>0}^{3}\rightarrow\mathbb{R}_{>0}$ be any positive
$1$-homogeneous function, $\rho(sx)=s\rho(x)$, and define
\[
\mathcal{C}_{\rho}=\{c\in\mathbb{R}_{>0}^{3}:\rho(c)=1\}.
\]
A map $f$ satisfies Eq.~\eqref{eq:supp-equiv} if and only if there exists a
map $\Psi:\mathcal{C}_{\rho}\rightarrow\mathbb{R}^{d}$ such that
\begin{equation}
f(x)=\alpha\log_2\rho(x)u+
\Psi\!\left(\frac{x}{\rho(x)}\right).
\label{eq:supp-decomp}
\end{equation}
This characterization alone does not assert injectivity; transversality and
injectivity of the chromatic component are additional requirements.

\paragraph{Proof.}
Assume Eq.~\eqref{eq:supp-equiv}. For $c\in\mathcal{C}_{\rho}$, define
$\Psi(c)=f(c)$. For arbitrary $x$, set $c=x/\rho(x)$. Homogeneity gives
$\rho(c)=1$ and $x=\rho(x)c$. Hence
\[
f(x)=f(c)+\alpha\log_2\rho(x)u
=\Psi(c)+\alpha\log_2\rho(x)u,
\]
which is Eq.~\eqref{eq:supp-decomp}. Conversely, homogeneity gives
$\rho(sx)=s\rho(x)$ and
$sx/\rho(sx)=x/\rho(x)$. Substitution into Eq.~\eqref{eq:supp-decomp} yields
\[
f(sx)=
\alpha\log_2(s\rho(x))u+
\Psi\!\left(\frac{x}{\rho(x)}\right)
=f(x)+\alpha\log_2(s)u.
\]
\(\square\)

\subsection{The Realized Encoding}

The deployed \textsc{EquivFlow} encoding is
\begin{equation}
f(x)=B
\begin{bmatrix}
\ell_{\tau}(Y(x))\\ u'(x)\\v'(x)
\end{bmatrix}+b,
\label{eq:supp-realized}
\end{equation}
where $Y,u',v'$ are Rec.\ 709 luminance and CIE 1976 chromaticity. The
toe-limited log luminance is
\begin{equation}
\ell_{\tau}(Y)=
\begin{cases}
\log_2Y,&Y\geq\tau,\\
\log_2\tau+(Y/\tau-1)/\ln 2,&0\leq Y<\tau.
\end{cases}
\label{eq:supp-toe}
\end{equation}
The frozen numerical values are
\[
\resizebox{0.98\columnwidth}{!}{$
B=\begin{bmatrix}
0.0512156526& 1.0503072513& 0.1687983249\\
0.0553585156&-0.4965997474& 0.3491496748\\
0.0499266056& 0.0399719486&-1.0148571856
\end{bmatrix}
$},
\]
\[
\resizebox{0.98\columnwidth}{!}{$
b=\begin{bmatrix}
0.4599701209&0.6956425436&1.2171179097
\end{bmatrix}^{\mathsf T},
\quad
\tau=0.0010543414867
$}.
\]

\paragraph{Analytic inverse and transversality.}
The scalar function $\ell_{\tau}$ is strictly increasing on $Y\geq0$, with
range $[t_{\min},\infty)$, where
$t_{\min}=\log_2\tau-1/\ln2$. Its inverse is
\begin{equation}
\ell_{\tau}^{-1}(t)=
\begin{cases}
2^t,&t\geq\log_2\tau,\\
\tau[1+\ln2\,(t-\log_2\tau)],
&t_{\min}\leq t<\log_2\tau.
\end{cases}
\label{eq:supp-toe-inverse}
\end{equation}
Moreover, the selected $B$ is nonsingular: it has rank 3,
$\det B=0.10697$, and spectral condition number $13.77$. Thus, for a code
$z$ in the image of Eq.~\eqref{eq:supp-realized}, one first computes
$q=B^{-1}(z-b)$, recovers $Y=\ell_{\tau}^{-1}(q_1)$, and then converts
$(Y,q_2,q_3)=(Y,u',v')$ uniquely back to linear Rec.\ 709 RGB. The standard
$Yu'v'$ inverse is unique on the admissible positive-color domain
($Y>0,v'>0$), and the Rec.\ 709 RGB--XYZ matrix is nonsingular. Therefore
Eq.~\eqref{eq:supp-realized} is injective on that domain.

This argument does not require the exposure and chromatic code directions to
be orthogonal. Define the dual functional
\[
\lambda^{\mathsf T}=e_1^{\mathsf T}B^{-1}.
\]
Then $\lambda^{\mathsf T}Be_1=1$ and
$\lambda^{\mathsf T}Be_2=\lambda^{\mathsf T}Be_3=0$. Consequently,
$\lambda^{\mathsf T}(z-b)$ recovers the luminance coordinate, while the
chromatic plane $\operatorname{span}(Be_2,Be_3)$ lies in
$\ker\lambda^{\mathsf T}$. This is the non-orthogonal direct-sum
transversality realized by the fitted matrix.

\paragraph{Exactness domain.}
Above the toe, scaling RGB by $s$ leaves $(u',v')$ unchanged and increments
the first coordinate by $\log_2s$. Hence Eq.~\eqref{eq:supp-equiv} holds
whenever both $Y$ and $sY$ are at least $\tau$, with
$\alpha=\lVert Be_1\rVert_2$ and $u=Be_1/\lVert Be_1\rVert_2$.
The linear toe is deliberately finite at black and is not exactly
log-equivariant when either endpoint lies below $\tau$.

The implementation floors RGB components at $10^{-8}$ before computing
$u'v'$, clamps inverse RGB to the nonnegative orthant, and clamps encoded
codes to $[0,1]$ at the VAE interface before mapping them to $[-1,1]$.
Accordingly, analytic injectivity and equivariance refer to the mathematical
encoding on the interior where these guards and code clipping are inactive;
they are not claims of global invertibility or exact equivariance for the
clipped VAE pipeline.

\section{Encoding Fitting and Selection}

\paragraph{Relation to the main paper.}
The main paper describes encoding fitting at the interface level: the
closed-form parameters of its Eq.~(3) are fitted by frozen-VAE round-trip
measurements, and Figure~2 (Step~1) depicts that fitting as a single block.
Concretely, this block is realized in two sub-stages. Stage~1 optimizes an auxiliary intermediate map through the frozen
VAE by gradient descent; its round-trip component corresponds to main-paper
Eq.~(7) (see the normalization note below), and the main paper's code-range
requirement enters as the range penalty of the stage-1 objective,
Eq.~\eqref{eq:supp-fit-loss} below.
Stage~2 produces the deployed encoding in closed form: the stage-1 targets
are projected into the affine family of main-paper Eq.~(3) by least squares
(affine distillation), and one displacement scale is then chosen by the
prespecified selection gates. The deployed parameters are therefore obtained
as follows: $B$ and $b$ from the closed-form projection, the toe $\tau$
carried over from the stage-1 map and frozen, and the first-column scale
from the frozen selection rule. Gradient optimization occurs only inside
stage~1; the stage-1 map is scaffolding and is discarded after
distillation.

\paragraph{Calibration data.}
The encoding was fitted once through the frozen LTX video VAE. The fixed
calibration pool contains 768 crops: four crops from each of 192 temporal
windows. Each crop has 9 frames at $256\times256$. With seed 0, saturation is
sampled uniformly from $[.55,.80]$ and, with probability .5, hue is perturbed
uniformly within $[-40^\circ,40^\circ]$.

\paragraph{Stage 1: VAE-aware fitting targets.}
Stage 1 optimizes an intermediate map $g$ only to provide targets
for affine distillation.
Concretely, $g$ is the fixed-form analytic map
\[
g(x)=s_{\mathrm{out}}\odot
M_2\{a\odot\operatorname{asinh}(b\odot M_1x)+d\}+b_{\mathrm{out}},
\]
where $M_i=I+\Delta M_i$;
its active tensors comprise the two $3{\times}3$ matrices; the
three-channel $a,b,d,s_{\mathrm{out}},b_{\mathrm{out}}$; an unconstrained
three-vector normalized to the unit gauge $u$; and one log-scale whose
exponential is $\alpha$ --- 37 scalar parameters, of which $u$ and $\alpha$
serve only the structural terms below.
Its internal parameterization is not part of the
deployed encoding. The final affine map instead obtains its realized exposure
flow directly from $Be_1$.

The frozen VAE is kept in the computation graph:
codes are clamped to $[0,1]$, mapped to $[-1,1]$, encoded and decoded in
32-bit floating point, and mapped back before analytic inversion. VAE
parameters receive no gradient. One exposure from
$\mathcal E_{\rm fit}=\{0,-4,-2,+2,+4\}$ EV is sampled uniformly at each
step.

For a VAE-round-tripped reconstruction $\hat x$, the intermediate fitting objective is
\begin{equation}
\mathcal{L}_{\mathrm{fit}}=
\underbrace{\mathcal{L}_{\log}+\mathcal{L}_{uv}}_{\mathcal L_{\mathrm{rt}}}
+\underbrace{\mathcal{L}_{\mathrm{eq}}+0.25\mathcal{L}_{\mathrm{amp}}}_{\mathcal L_{\mathrm{struct}}}
+10\mathcal{L}_{\mathrm{range}}.
\label{eq:supp-fit-loss}
\end{equation}
The three groups respectively measure VAE round-trip fidelity, regularize the
exposure--chroma structure, and keep codes inside the useful VAE input range.
The weights of $\mathcal{L}_{\log}$, $\mathcal{L}_{uv}$, and
$\mathcal{L}_{\mathrm{eq}}$ are one. Specifically,
\[
\mathcal{L}_{\log}
=\operatorname{mean}\lVert\log(1+\hat x)-\log(1+x)\rVert_2^2,
\]
and $\mathcal{L}_{uv}$ is the mean Euclidean $u'v'$ error over pixels with
ground-truth $Y\geq10^{-3}$. Thus the compact notation in the main paper
denotes a masked per-pixel mean, rather than an unnormalized tensor norm.
For $e\in\mathcal{E}_{\mathrm{fit}}$, $\mathcal{L}_{\mathrm{eq}}$ penalizes
\[
\operatorname{mean}_{m}
\lVert g(2^ex)-g(x)-e\alpha u\rVert_2^2,
\]
where \(g\) is the intermediate fitting map, \(u\) is its learned unit gauge, and
\(\alpha>0\) is its learned displacement scale. The term
$\mathcal{L}_{\mathrm{amp}}$ penalizes the change with exposure of the
code distance between an image and its chromatically perturbed counterpart.
The same valid-luminance mask $m=\mathbf{1}[Y\geq10^{-3}]$ is used.
Finally,
\[
\begin{aligned}
\mathcal{L}_{\mathrm{range}}=\operatorname{mean}\{&
\operatorname{ReLU}(g(x)-0.95)^2\\
&+\operatorname{ReLU}(0.05-g(x))^2\}.
\end{aligned}
\]
We use AdamW for 15,000 steps with learning rate $2\times10^{-3}$, weight
decay $10^{-4}$, batch size 4, cosine decay to zero, unit gradient clipping,
and seed 0.

\paragraph{Stage 2: affine distillation (closed form).}
The distillation targets are the code vectors $g(x)$ of the converged
stage-1 map on 120,000 RGB samples at $\{0,-2,-4,-6,-8\}$ EV (600,000
fitting rows); Eq.~\eqref{eq:supp-realized} is fitted to these targets by
least squares. The $-6$ and $-8$ EV rows evaluate the fixed target map outside its
optimization ladder; they do not extend the first-stage optimization. The toe is frozen,
the first column of $B$ is swept over
$c\in\{.5,.75,1,1.5,2,2.5,3\}$, and the bias is recentered using 40,000
additional samples.

Since exposure moves the code along $Be_1$ with $\alpha=\lVert Be_1\rVert_2$
(Eq.~\eqref{eq:supp-equiv}), scaling that column by $c$ rescales the per-stop
displacement relative to the fixed chromatic columns, trading round-trip
reconstruction against deep-exposure chromatic separation; Table~\ref{tab:supp-selection} lists the gate
thresholds and the two passing candidates of this sweep. The selected $c$
is not a separate deployed parameter: it is absorbed into the frozen matrix
of Eq.~\eqref{eq:supp-realized}, whose printed first column already
contains it.

Selection is performed before downstream task evaluation. Reconstruction
uses a fixed 16-clip encoding-fit validation panel.
OOD survival is the ratio
of held-out chromaticity \(R^2\) at \(-8\) EV to that at EV 0 on a fixed
12-clip ARRI--UPIQ OOD survival panel. Grid identification is the four-fold texture-held-out
classifier defined below. The projection score is a group-held-out linear
probe on absolute top-24-PCA perturbation coordinates; the separation AUC
uses the fraction of perturbation energy assigned to independently estimated
luminance and chroma subspaces.

\begin{table*}[t]
\centering
\small
\setlength{\tabcolsep}{6.0pt}
\begin{tabular}{@{}lrrrrrr@{}}
\toprule
candidate & recon. $\uparrow$ & OOD surv. $\uparrow$ & grid$_{-8}\uparrow$ &
exposure $R^2\uparrow$ & projection $\uparrow$ & separation AUC $\uparrow$ \\
\midrule
gate & 35.99 & $0.8\times$ fit ref. & .90 & .80 & .89 & .85 \\
$c=2.5$ & 36.08 & 1.092 & .979 & .869 & .985 & 1.000 \\
$c=3$ & \textbf{36.28} & 1.102 & .979 & .864 & .999 & 1.000 \\
\bottomrule
\end{tabular}
\caption{Encoding-selection gates and the two passing candidates. Candidates
differ only in the exposure-column scale $c$ (Stage~2); the predefined
highest-reconstruction rule selects $c=3$. These are selection-panel
scores; the separate main-paper probe panel gives 34.38 dB for the selected
encoding.}
\label{tab:supp-selection}
\end{table*}

No downstream SDR-to-HDR prediction or task metric enters encoding selection;
the selection and main-paper reconstruction values estimate different panels.

Encoding selection predates the downstream subset aggregation, but two of the 12
OOD-survival-panel clips also occur in the frozen ARRI subset of the test set. Thus no
downstream output or metric is used for selection, but encoding selection and downstream
evaluation are not strictly image-disjoint. We treat this gate as design selection and do
not claim an untouched-image test for representation selection.

\begin{table*}[t]
\centering
\small
\setlength{\tabcolsep}{5.0pt}
\begin{tabular}{@{}crrrrl@{}}
\toprule
$\mathcal L_{uv}$ & PU21-PSNR $\uparrow$ & $\Delta u'v'\downarrow$ &
chroma L1 $\downarrow$ & $\Delta C_TC_P\downarrow$ & selection \\
\midrule
on  & \textbf{35.95} & \textbf{.0140} & \textbf{.0431} & \textbf{6.37} & pass; \textsc{EquivFlow} \\
off & 34.22 & .0219 & .0582 & 10.28 & fail; waived \\
\bottomrule
\end{tabular}
\caption{Chromatic-loss ablation under the training Reinhard curve on the full
HDRTV4K--UPIQ evaluation pool (100 clips, 58 parents). Entries are
parent-weighted descriptive means from separately fitted encodings and
separately trained adapters. The failed encoding is not admissible; its
downstream result is reported only to diagnose the removed term.}
\label{tab:supp-loss-ablation}
\end{table*}

\subsection{Chromatic-Loss Ablation}

We rerun both fitting stages (stage-1 optimization and stage-2 affine
distillation), the scale sweep, the frozen selection rule, and 6,500-step
adapter training with the chromatic round-trip
term $\mathcal L_{uv}$ removed.
All other loss terms and settings are fixed.
This fit produces no candidate that passes the selection rule; for diagnostic
downstream readout only, its best-reconstruction candidate is exported with
the gate explicitly waived.

Removing $\mathcal L_{uv}$ costs 1.73 dB PU21 and increases mean
$\Delta u'v'$ from .0140 to .0219. The ablation identifies
$\mathcal L_{uv}$ as a load-bearing term in the \textsc{EquivFlow} fitting objective.

\section{SDR-to-HDR System Details}

\subsection{Data Construction}

\paragraph{HDR targets and split.}
The training corpus contains 5,780 crops from 1,015 parent sequences:
2,580 crops from 215 temporal windows extracted from 94 \emph{Tears of
Steel} shots, and 3,200 crops from 800 PolyHaven clips. The
parent-disjoint split contains 5,388 training, 324 validation, and 68 test
crops. Each clip is normalized by mapping the 75th percentile of its
per-frame 99th-percentile luma to unity.

The two sources differ in provenance. PolyHaven renders are scene-referred
and retain highlights above reference white. \emph{Tears of Steel} is a
finished 16-bit display grade that we linearize with the inverse sRGB
transfer function; its grade has already compressed highlight information.
We therefore refer to the combined data as linear Rec.\ 709 targets, not
uniformly as scene-linear radiance.

\paragraph{SDR generation.}
Each target has four offline SDR variants. Variant 0 is a clean lossless
FFV1 render. Variants 1--3 use independently sampled degradation parameters.
The loader selects the clean variant with probability $0.25$ and otherwise
selects uniformly among the three degraded variants, so each stored variant
has marginal probability $0.25$. The variant generator uses seed
$42+\text{sample index}$. No online exposure, contrast, blur, or codec
augmentation is applied during adapter training.

All training variants use extended Reinhard with white point $w=4$ and no
additional exposure offset. For linear luminance $L$,
\begin{align}
L_t&=\frac{L(1+L/w^2)}{1+L},\nonumber\\
L&=\frac{w^2}{2}\left[
(L_t-1)+\sqrt{(1-L_t)^2+\frac{4L_t}{w^2}}
\right].
\label{eq:supp-reinhard}
\end{align}
The luminance ratio is applied to RGB, followed by per-channel clipping,
sRGB OETF, and 8-bit quantization. The analytic inverse in
Eq.~\eqref{eq:supp-reinhard} inverts the unclipped continuous luminance
curve; clipping and quantization remain information-losing.

For degraded variants, sRGB contrast about 0.5 is sampled uniformly from
$[0.85,1.25]$. Gaussian blur with $\sigma\sim U[1,3]$ is applied selectively
to highlights above 0.98 and/or shadows below 0.02, each independently with
probability 0.7. H.264 encoding uses \texttt{yuv420p}, integer CRF sampled
from 18--32, and the \texttt{medium} preset.

\paragraph{Unseen tone curve.}
The ACES-like evaluation condition applies
\begin{equation}
A(y)=\operatorname{clip}\!\left(
\frac{y(2.51y+0.03)}{y(2.43y+0.59)+0.14},0,1\right)
\label{eq:supp-aces}
\end{equation}
as a luminance-preserving RGB ratio, followed by the same sRGB transfer and
8-bit quantization. This curve is not used to train either internal arm.

\subsection{Adapter Training and Inference}

The base model is the LTX-2.3 22B checkpoint
\texttt{ltx-2.3-22b-dev.safetensors}; its transformer, VAE, and text encoder
remain frozen. IC-LoRA reference tokens use clean ($t=0$) latents,
participate in bidirectional attention, and are excluded from the denoising
loss. Rank-32 modules use scale 32 and dropout 0. They are inserted into the
$q$, $k$, $v$, and output projections of \texttt{attn1} and \texttt{attn2},
and into \texttt{ff.net.0.proj} and \texttt{ff.net.2}.

Each arm is trained once for a fixed 6,500-step comparison budget, rather
than a validation-selected checkpoint. We use AdamW at learning rate
$2\times10^{-4}$ with unit gradient clipping and a linear factor
$1\!\rightarrow\!.1$ over a 100,000-step horizon. Training uses four NVIDIA
H100 80GB GPUs, per-GPU batch size 1, accumulation 1 (effective batch 4),
BF16, gradient checkpointing, and seed 42. Noise follows a shifted
logit-normal distribution with standard deviation 1, $\epsilon=10^{-3}$, a
.1 uniform mixture, and the model's sequence-dependent shift.

The positive prompt is ``A high dynamic range cinematic scene with bright
highlights and deep shadows.''; the negative prompt is ``worst quality,
inconsistent motion, blurry, jittery, distorted.'' Inference uses seed 42,
30 steps, guidance 4, and spatiotemporal guidance 1 at block 28 in
\texttt{stg\_v} mode. All settings other than the conditioning contract and
encoding are shared: EquivFlow conditions on the analytically
inverse-tone-mapped proxy, whereas LogC3 uses the display-referred SDR
directly.

\section{Evaluation Definitions}

\paragraph{Common output domain.}
All outputs are decoded to linear Rec.\ 709 before evaluation. PU21 uses
$100$ nits per normalized linear unit, clamps absolute luminance to
$[0.005,10,000]$ nits, applies the banding--glare PU21 encoding normalized
to unit peak, and reports PSNR with peak 1.

For chromatic \(\Delta C_TC_P\), Rec.\ 709 is converted to Rec.\ 2020 and
then to PQ ICtCp using the 10,000-nit convention. We use the ITP scaling
$(I,T,P)=(I,0.5C_T,C_P)$ and report the mean
\[
720\sqrt{(0.5\Delta C_T)^2+(\Delta C_P)^2}.
\]
This omits the intensity difference by construction.

\paragraph{Luminance-independent chroma metrics.}
Let
$r(x)=x/(\sum_c x_c+\epsilon)$, with $\epsilon=10^{-4}$ used only
to avoid division by zero. Chroma L1 is
$\operatorname{mean}|r(\hat x)-r(x)|$. Mean $\Delta u'v'$ is the Euclidean
CIE 1976 chromaticity error over ground-truth pixels with $Y\geq10^{-3}$.
LF drift first applies a spatial Gaussian with $\sigma=8$ to the
chromaticity fields and measures the same masked error. Range drift further
restricts evaluation to pixels whose SDR input is neither clipped nor null:
all channels are below $254/255$ and the maximum channel exceeds $3/255$.
The null region is replaced by the target field before Gaussian filtering
to prevent invalid boundaries from leaking into the range mask.

\paragraph{Main-paper frozen-VAE probe.}
Table 1 in the main paper uses no diffusion training. Its reconstruction
column is the mean per-clip log-RGB PSNR after a frozen-VAE round trip on 16
deterministically sampled held-out validation clips (a panel distinct from
the encoding-selection panel of Table~\ref{tab:supp-selection}).
For the chromatic columns,
12 deterministic center-cropped clips from the ARRI--UPIQ OOD pool are
evaluated at \(\{0,-2,-4,-6,-8,-10\}\) EV. Only latent tokens whose EV-0
ground-truth luminance is at least \(10^{-3}\) are used. At every exposure
and for each encoding separately, a \texttt{StandardScaler} followed by
ridge regression with penalty 1 is refitted to predict \(u'\) and \(v'\).
Five-fold \texttt{GroupKFold} by parent clip supplies held-out predictions;
the reported \(R^2_{u'v'}\) is the mean of the two coordinate scores and
\(\mathrm{surv}_{-k}=R^2(-k)/R^2(0)\). Because held-out \(R^2\) is unbounded
below and the numerator can exceed its EV-0 value, survival may be negative
or exceed one. It measures relative linear recoverability, not chroma-energy
amplification.

\paragraph{Cross-VAE probe.}
The structural-portability test applies the same fixed encoding to each
frozen host VAE over the exposure ladder
\(\{0,-2,-4,-6,-8\}\). Its chroma grid comprises 12 chromaticities (two
radii, 0.03 and 0.06, at six equally spaced hue angles about D65) rendered
with four luminance textures at base luminance 0.45. Grid identification is
12-way accuracy; chance is $1/12$. At each host and exposure, a
\texttt{StandardScaler} plus logistic regression (\(C=1\)) is evaluated by
four-fold \texttt{GroupKFold}, holding out one luminance texture per fold.
Thus the table's grid readout is host- and exposure-specific.

For the exposure readout, a host-specific direction is the first right
singular vector of all nonzero-EV latent differences. The scalar
\(\beta(e)\) is their mean projection on that direction, and
\(R^2_{\beta}\) is the no-intercept linear fit of \(\beta(e)\) to EV. The
reported \(\Delta u'v'_{-8}\) is the encoding round-trip chromaticity error at
$-8$ EV. The prespecified non-collapse check uses 16 fixed clips (15 with a
nonempty EV-0 luminance mask) and requires each foreign host's round-trip
\(\Delta u'v'\) to be no more than twice the LTX value at both \(-6\) and
\(-8\) EV. This is a within-subject collapse check, not a comparative
superiority test.

\paragraph{EV-0-frozen readout.}
The main paper notes that an EV-0-frozen readout degrades along each host's
exposure trajectory. This diagnostic fits the chroma-grid classifier at
EV~0 and applies it unchanged at deeper exposures, instead of refitting per
exposure as in the grid column of main-paper Table~4. Under the frozen
readout, $-8$~EV grid accuracy falls to .208 (LTX), .083 (SD~2.1, chance
level), .167 (Wan~2.1), and .250 (Wan~2.2) from EV-0 values of .979, .667,
.958, and .979, while the per-exposure refitted readout remains high
(main-paper Table~4). The latent coordinates therefore rotate or rescale
along each host's exposure trajectory even though exposure structure and
chroma separability persist; the portability claim concerns the latter,
not shared latent coordinates.

\begin{table*}[t]
\centering
\small
\setlength{\tabcolsep}{6.0pt}
\begin{tabular}{@{}lrrr@{\hspace{20pt}}rrr@{}}
\toprule
& \multicolumn{3}{c@{\hspace{20pt}}}{Reinhard} & \multicolumn{3}{c}{ACES-like} \\
\cmidrule(lr{20pt}){2-4}\cmidrule(l){5-7}
metric & \(\Delta\) & 95\% CI & \(p\) value & \(\Delta\) & 95\% CI & \(p\) value \\
\midrule
chroma L1 & \(-0.01083\) & \([-0.01565,\,-0.00598]\) & \(1.61{\times}10^{-6}\) & \(-0.01317\) & \([-0.01834,\,-0.00799]\) & \(2.47{\times}10^{-7}\) \\
\(\Delta u'v'\) & \(-0.00330\) & \([-0.00451,\,-0.00209]\) & \(1.61{\times}10^{-6}\) & \(-0.00769\) & \([-0.00976,\,-0.00577]\) & \(6.04{\times}10^{-10}\) \\
\(\Delta C_TC_P\) & \(-0.998\) & \([-1.388,\,-0.614]\) & \(4.83{\times}10^{-6}\) & \(-1.489\) & \([-1.926,\,-1.065]\) & \(1.65{\times}10^{-7}\) \\
\addlinespace[3pt]
LF drift\textsuperscript{\(\dagger\)} & \(-0.00417\) & \([-0.00572,\,-0.00265]\) & \(4.20{\times}10^{-7}\) & \(-0.00768\) & \([-0.00968,\,-0.00583]\) & \(1.49{\times}10^{-10}\) \\
range drift\textsuperscript{\(\dagger\)} & \(-0.00407\) & \([-0.00557,\,-0.00266]\) & \(1.37{\times}10^{-7}\) & \(-0.00807\) & \([-0.01010,\,-0.00613]\) & \(6.55{\times}10^{-11}\) \\
\bottomrule
\end{tabular}
\caption{Parent-paired chroma differences
(\textsc{EquivFlow}\(-\)LogC3) on the pooled HDRTV4K--UPIQ panel (58 parent
scenes); negative values favor \textsc{EquivFlow}, and every bootstrap
interval excludes zero. The three primary rows report Holm-corrected
\(p_{\mathrm H}\) within each curve's three-metric family.
\textsuperscript{\(\dagger\)}Nested \(\Delta u'v'\) variants are descriptive
and report uncorrected \(p\).}
\label{tab:sup-chroma-stats}
\end{table*}

\paragraph{LEDiff radiometric alignment.}
LEDiff is run with its official preprocessing and checkpoint. A single
method-level scalar gain is fitted once on 24 fixed, parent-distinct training
clips and then held fixed for every evaluation set. With luminance \(y\), the
gain is
\[
\begin{aligned}
g_{\rm LEDiff}
&=\exp\!\left(\operatorname{mean}_{y_{\rm gt}>10^{-4}}\delta\right)
=0.0294942,\\
\delta&=\log(y_{\rm gt}+10^{-4})-\log(y_{\rm pred}+10^{-4}).
\end{aligned}
\]
No test image, test ground truth, or per-clip gain enters this fit.

\paragraph{Qualitative selection and display.}
For each evaluation set, the main-paper figure selects the evaluated crop
with the largest \textsc{EquivFlow}--LumiVid
\(\Delta E_{\mathrm{ITP}}\) margin; it therefore shows selected successes,
not representative or median cases. ARRI and HDRTV4K are displayed in the
full frame containing that crop, whereas UPIQ contains one evaluated crop.
The inset is chosen as the darkest chroma-carrying target region that remains
inside one reconstructed crop. A row-specific display gain is applied
identically to every column and affects no reported metric. For HDRTV4K, only
the evaluated quadrant contributes to quantitative results; the other
quadrants are display context.

\section{Statistical Protocol and Full Results}

\paragraph{Frozen evaluation subset.}
The frozen 102-clip test subset of main-paper Table~2 was fixed before any
subset aggregation was computed.
Table~\ref{tab:supp-testset}
records the per-set selection rule.

\begin{table}[t]
\centering
\small
\setlength{\tabcolsep}{3.4pt}
\begin{tabular}{@{}llrr@{}}
\toprule
set & selection & clips & parents \\
\midrule
ARRI & first 40 clips & 40 & 10 \\
HDRTV4K & first 40 clips & 40 & 19 \\
UPIQ & all parents & 22 & 22 \\
\bottomrule
\end{tabular}
\caption{Frozen evaluation subset. ``First'' refers to each set's fixed
evaluation ordering; the ACES-like condition uses the same clips.}
\label{tab:supp-testset}
\end{table}

\paragraph{Aggregation and inference.}
Metrics are first averaged within each parent and then averaged with equal
weight across parents. The chroma-decomposition panel uses the full
HDRTV4K--UPIQ pool of 58 parents (36 HDRTV4K and 22 UPIQ).
Confidence intervals are nonparametric parent-cluster bootstrap intervals
from 10,000 resamples with seed 42. Pairwise tests are two-sided,
parent-level paired Wilcoxon signed-rank tests. For the dataset-wise table,
Holm correction is applied within each method pair and curve across three
datasets and two primary metrics. For the chroma panel, correction is
applied within each curve across its three primary metrics. There is one
trained model per internal arm; bootstrap intervals characterize
parent-level sampling uncertainty rather than variation across training
runs.

\paragraph{Dataset-wise internal comparison.}
Table~\ref{tab:sup-dataset-stats} supplies the parent-level tests behind the
dataset-wise statements in the main paper. Deltas are
\textsc{EquivFlow}\(-\)LogC3, so positive PU21-PSNR and negative
\(\Delta C_TC_P\) favor \textsc{EquivFlow}. The UPIQ chromatic difference is
significant under the training curve; HDRTV4K and UPIQ are significant under
the unseen curve. ARRI does not separate under either curve, and HDRTV4K
does not separate under the training curve.

\begin{table}[t]
\centering
\small
\setlength{\tabcolsep}{2.4pt}
\begin{tabular}{@{}llrrrr@{}}
\toprule
curve & set & \(\Delta\)PU21 & \(p_{\mathrm H}\) &
\(\Delta C_TC_P\) & \(p_{\mathrm H}\) \\
\midrule
reinhard & ARRI     & \(+0.294\) & .465 & \(-0.013\) & .492 \\
         & HDRTV4K  & \(+1.684\) & .312 & \(-1.066\) & .197 \\
         & UPIQ     & \(+2.485\) & .00128 & \(-0.917\) & .00128 \\
\midrule
aces-like & ARRI    & \(-0.216\) & .465 & \(-0.259\) & .465 \\
          & HDRTV4K & \(-1.282\) & .00134 & \(-1.718\) & .00508 \\
          & UPIQ    & \(-1.024\) & .00134 & \(-1.327\) & .000157 \\
\bottomrule
\end{tabular}
\caption{Internal pair on the frozen test subset. \(p_{\mathrm H}\) is the
two-sided paired Wilcoxon value after Holm correction within the stated
six-test family.}
\label{tab:sup-dataset-stats}
\end{table}

\paragraph{Chroma-panel statistics.}
Table~\ref{tab:sup-chroma-stats} gives the paired differences underlying the
main paper's chroma panel for the fixed 6,500-step \textsc{EquivFlow}
checkpoint. Units are parent
scenes ($n=58$); each entry is
\textsc{EquivFlow}\(-\)LogC3 with a 10,000-sample bootstrap 95\% interval.
The three primary metrics use Holm-corrected \(p_{\mathrm H}\). The two
nested filtered forms of $\Delta u'v'$ are descriptive, use uncorrected
$p$, and lie outside the correction family.

\paragraph{Interpretive boundary.}
Under matched Reinhard, EquivFlow has higher aggregate-fidelity point
estimates on the frozen subset. That aggregate advantage does not persist
under the ACES-like swap. The result that remains consistent across the two
tested curves is narrower: the full EquivFlow system has lower
luminance-independent chromaticity error than the same-backbone LogC3
system. Because the arms also differ in reference reconstruction, these
results attribute the effect to the complete conditioning system, not to the
encoding alone.

\end{document}